\documentclass{egpubl}
\usepackage{eurovis2022}
\usepackage{enumitem}

\EuroVisShort  

\usepackage[T1]{fontenc}
\usepackage{dfadobe}  

\usepackage{cite}  
\BibtexOrBiblatex
\electronicVersion
\PrintedOrElectronic
\ifpdf \usepackage[pdftex]{graphicx} \pdfcompresslevel=9
\else \usepackage[dvips]{graphicx} \fi

\usepackage{egweblnk}

\title[\textit{Data+Shift}: Supporting visual investigation of data distribution shifts by data scientists]{\textit{Data+Shift}: Supporting visual investigation of data distribution shifts by data scientists}

\author[J. Palmeiro et al.]
{
    \parbox{\textwidth}{\centering João Palmeiro, Beatriz Malveiro, Rita Costa, David Polido, Ricardo Moreira, and Pedro Bizarro}\\
    {\parbox{\textwidth}{\centering Feedzai}}
}

\begin{document}
\teaser{
    \includegraphics[width=\linewidth]{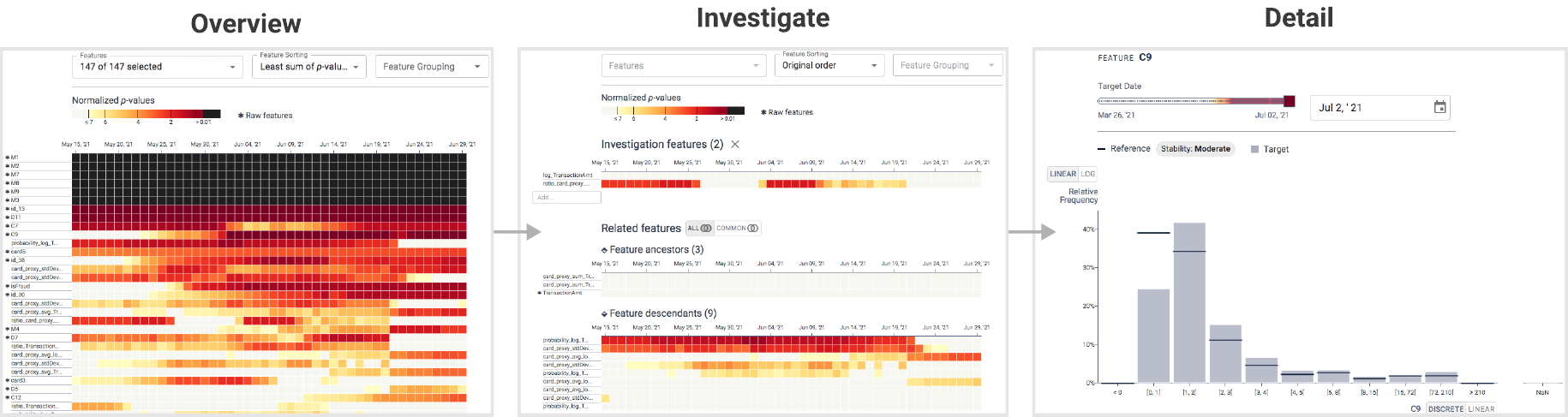} 
    \centering
    \caption{Views supported by \textit{Data+Shift} to assist data scientists when investigating data distribution shift. The overview (left) shows how much the distribution of each feature has changed over time. The investigation view (center) enables exploration of related features to better understand sources and the impact of the shift. In addition to the selected feature(s), the related features are shown grouped in ancestors and descendants. The detail view (right) shows how the distribution of a single feature has changed over time.}
    \label{fig:teaser}
}

\maketitle

\begin{abstract}
   Machine learning on data streams is increasingly more present in multiple domains. However, there is often data distribution shift that can lead machine learning models to make incorrect decisions. While there are automatic methods to detect when drift is happening, human analysis, often by data scientists, is essential to diagnose the causes of the problem and adjust the system. We propose \textit{Data+Shift}, a visual analytics tool to support data scientists in the task of investigating the underlying factors of shift in data features in the context of fraud detection. Design requirements were derived from interviews with data scientists. \textit{Data+Shift} is integrated with JupyterLab and can be used alongside other data science tools. We validated our approach with a think-aloud experiment where a data scientist used the tool for a fraud detection use case.\\


\end{abstract}

\section{Introduction}

Machine learning (ML) on real-time stream processing systems is increasingly present in multiple domains. These complex systems rely on a data stream that is assumed to follow the same distribution as previously seen. However, there is often data shift which impacts the system performance and leads to degradation over time~\cite{pinto2019automatic}.
In the fraud detection domain, data shift can result from multiple factors, such as seasonality, new buying patterns, or some technical issues that can corrupt the observed data. Depending on the underlying factor behind the drift, its effect can be gradual or sudden, and impact specific features of the data or all of them~\cite{DalPozzolo2018CreditCF}. In the data streams literature, these changes are known as covariate shift and concept drift. Several methods~\cite{pinto2019automatic, Bifet07learningfrom, 10.1007/s10994-012-5320-9, 10.1145/2939672.2939836} have been proposed to detect these changes. These automatic systems can signal data features that might be drifting and in which periods. However, while this automatic signal is important to kick-start an investigation, it is only the beginning. Human analysis, often by data scientists, is essential to diagnose what factors impact the system and then take appropriate measures to adjust it.

In this paper, we introduce a visual analytics tool to support data scientists in the tasks of investigating the underlying factors of drift in data features: \textit{Data+Shift}. Our tool relies on an algorithm that calculates if a feature's distribution has significantly diverged from a reference. This method uses hypothesis testing to check if the difference between distributions is significant. Our visual interface provides a way to analyze the \textit{p}-values of these tests across time for all features. The interactive visualization starts with an overview and supports the user throughout the investigation (Figure~\ref{fig:teaser}). \textit{Data+Shift} is directed at data scientists that use JupyterLab~\cite{jupyter} to perform data analysis and model development. To validate the use of \textit{Data+Shift} for data distribution shift exploration, we gathered feedback from an expert. We asked a data scientist to freely explore the tool in the context of ML for fraud detection, following a think-aloud protocol~\cite{think-aloud}.

\section{Related Work}

Visual analytics for data distribution shift (as part of ML systems) is an underexplored field~\cite{data-shift-explorer}. For this work, the following topics are considered: (1) interfaces for data distribution shift visualization; (2) heatmaps/matrix-based visualizations for high-dimensional data; (3) time series visualization.

For identification and analysis of data distribution shifts, DataShiftExplorer~\cite{data-shift-explorer} presents a feature-oriented interface designed from a previously defined design space. It's the closest work to our proposed solution. On the other hand, Clustergrammer~\cite{clustergrammer} is an interactive heatmap visualization, embeddable in Jupyter Notebook, prepared particularly for biological data. In terms of implementation, it's the closest solution to ours, albeit for a substantially different use case. In the open-source software landscape, some packages~\cite{evidently,mlrun} offer visualization capabilities for data distribution shift analysis.

We also looked at the literature on heatmaps for high-dimensional data to complement our design process and account for techniques that could inspire our work focused on data distribution shift. ContiMap~\cite{contimap} features a heatmap that, instead of rendering individual cells, sorts and groups them, producing a high-level visualization. Similarly, HiGlass~\cite{higlass} also provides a scalable heatmap-based interface for Genomics. In Compadre's heatmap~\cite{compadre}, each cell encodes a distance between the original data and/or low-dimensional projections. Furthermore, the top and left borders of the heatmap are used to stack the labels associated with each point. In H-Matrix~\cite{h-matrix}, a multi-part linguistic visualization, it's possible to find a heatmap whose cells follow a divergent color scale for an error metric. Responsive matrix cells~\cite{responsive-matrix-cells} offer local zoomable regions, where the user can check other relevant visualizations \textit{in-situ} (with different levels of detail).


Similarly, since the analysis and comparison of features over time is relevant to \textit{Data+Shift}, we also consider techniques for visualizing time series applied in different contexts. CloudLines~\cite{cloudlines} is a dynamic and space-efficient time series visualization, using an importance function to adapt as new data becomes available. In ConceptExplorer~\cite{concept-explorer}, an interface for detecting and analyzing time series drifts, two snapshots (before and after a drift, for example) can be compared in a heatmap whose halves are independent. In addition, each feature on the axes is discretized or categorized, allowing for a more granular analysis. Particularly to explore cyclical patterns in time series, Ceneda et al.~\cite{ceneda-et-al} enrich a spiral chart with user-adjustable settings and sensible suggestions for them. SAX Navigator~\cite{sax-navigator} is a hierarchical interface for analyzing and comparing time series. They are simplified and clustered, and the user has access to small heatmaps with summaries for these clusters.

None of these works provide an ML-oriented workflow for analyzing data distribution shifts with various levels of information and interactivity like \textit{Data+Shift}, integrated with JupyterLab and Python.

\section{\textit{Data+Shift}}


In this section, we describe the algorithm used by \textit{Data+Shift}, the user requirements derived from interviews, and the visual interface.

\subsection{Algorithm}

The algorithm used by the tool aims at detecting covariate drift by measuring the divergences of each feature's distribution over time. Our algorithm is most similar to~\cite{pinto2019automatic,10.1145/2939672.2939836}, which does not rely on labels. The algorithm works in two steps. It first starts by learning the historical behavior of the features in the dataset and then evaluates how much they diverge over time.  

Similar to how machine learning models are trained, the historical behavior is learned from a reference dataset. This can be mapped to the "train" stage of the usual ML model training pipeline. The reference dataset covers a time interval when the data is assumed to be stable and follow expected patterns. In this stage, for every feature, a reference histogram of the distribution in the full reference dataset is stored. To measure the relative stability of the dataset in this period, sampled windowed histograms are also computed and then compared with the reference histogram obtaining a distribution of divergences.

In the second stage, we evaluate the drifts over time on a second dataset. As an example, let us assume we run a daily analysis. Then, for each feature, we compare its histogram in each day of the new dataset with the reference histogram and calculate its divergence measure. However, we need to understand how significant this divergence is compared to the expected distribution of divergences obtained from the reference dataset. We perform a test for what would be the likelihood of observing such a divergence value under the hypothesis that the feature distribution remains similar to the reference. The \textit{p}-value of this test is then the probability that the feature keeps the same distribution as in the reference. To create an alarm for a single feature, we can define a significance level $\alpha$ and use that as a threshold. If the \textit{p}-value is less than the threshold, we consider that the feature is under drift, and an alarm for it is triggered. In order for all the \textit{p}-values to be comparable with the same threshold, these need to be normalized \cite{Holm1979ASS}. Hence, the normalized \textit{p}-values are no longer probabilities, as they can take any value greater than zero.

\subsection{User Requirements}





To help us with the design process, we conducted semi-structured interviews at Feedzai with four data scientists. From the interviews, we concluded that, in general, data scientists want to compare the data distributions of features considering the most recent time window with the training, i.e., historical datasets. They leverage charts and summary statistics via single-use Jupyter notebooks or spreadsheets. When starting the analysis, they prioritize the most important features for predicting the target variable, raw fields, and features obtained from data enrichers. In addition, there are two main triggers for a data scientist to investigate features: changes in model/system performance and deployments.

The main goal is to quickly generate hypotheses and decide the next steps to ensure that the performance of ML models is not affected by data distribution shifts and data quality issues. In fraud detection, these hypotheses often consider seasonality, customer behavior changes, or technical issues. These causes are associated with different patterns of data distribution shift. For example, technical issues often cause sudden changes, while changing behavior happens gradually. For \textit{Data+Shift} to support data scientists in the tasks of generating new hypotheses and validating them, we have established the following user requirements:

\begin{enumerate}[label=\textbf{R\arabic*}, ref=R\arabic*, leftmargin=*]
    \item Identify and compare data distribution shift patterns.\label{rPvalues}
    \item Identify relevant features and periods.\label{rIdentifyFeature}
    \item Compare related features.\label{rCompareFeature}
    \item Identify the main changing values in the distributions of the features.\label{rDistributions}
    \item Integrate with the current workflow of data scientists.\label{rWorkflow}
\end{enumerate}

Although the four data scientists work in the fraud detection domain, the elicited requirements appear to be generic. \textit{Data+Shift} requires further testing to validate its application to other use cases.


\subsection{Visual Interface}
\subsubsection{Overview and Investigation} \label{overview}

\begin{figure}
    \centering
	\includegraphics[width=0.5\textwidth]{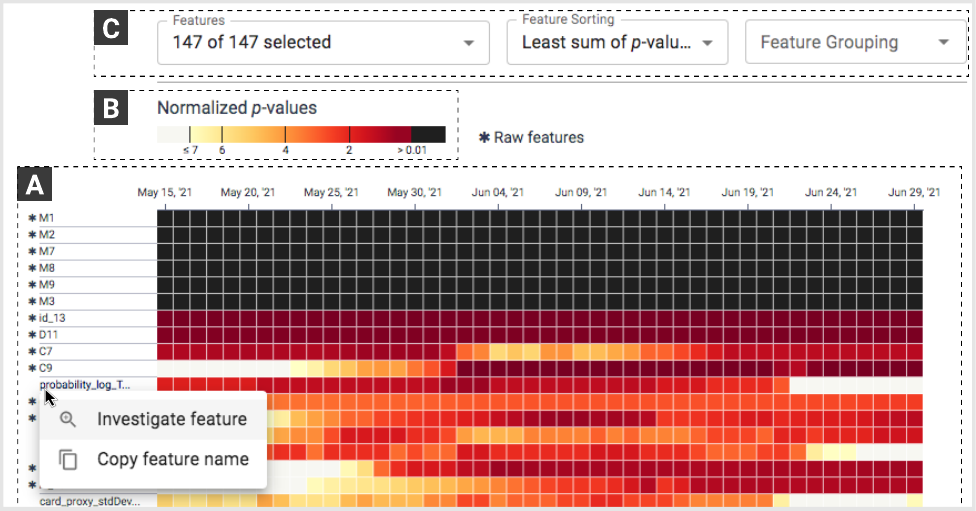}
	\caption{Overview screen. (A) Heatmap of the \textit{p}-values for all features. The color encoding is described in the legend (B). A toolbar (C) allows the user to interact with the heatmap.}
	\label{fig:overview}
\end{figure}

The overview visualization shows how the features diverged from their reference over time (\ref{rPvalues}, \ref{rIdentifyFeature}). The main element of the application is the heatmap (Figure~\ref{fig:overview}.A), where each cell color represents the \textit{p}-value for a given feature at a time instance.

\begin{itemize}
    \item \textbf{Tripartite color scale:} The legend on top of the heatmap describes the color encoding, which uses a tripartite scale (Figure~\ref{fig:overview}.B). Any \textit{p}-value above the analysis threshold will be colored in a light gray, a value lower than the alert threshold will be black, and a value in between these thresholds will be colored according to a gradient from light yellow to dark red. The analysis threshold serves to focus on a shorter, more relevant range of \textit{p}-values.
    \item \textbf{Toolbar:} It is possible to interact with the layout of the heatmap using the toolbar at the top (Figure~\ref{fig:overview}.C). It enables three distinct operations over the heatmap:
    
    \begin{itemize}
        \item \textbf{Selection} allows the user to search, select, and remove specific features from the view.
        \item \textbf{Sorting} allows the user to change the vertical ordering of the features to any of the four different sorting options: \textit{Original}, \textit{Alphabetical}, \textit{Most alarms}, and \textit{Least sum of \textit{p}-values}.
        \item \textbf{Grouping} allows for a vertical split of the heatmap, grouping features by a specific attribute (e.g., \textit{Raw / Engineered} breakdown).
    \end{itemize}
\end{itemize}


It is possible to investigate a feature in more detail using the \textbf{investigation view} (Figure~\ref{fig:teaser}, center). This view can be triggered from any single feature. When right-clicking on a feature name on the heatmap Y-axis, a context menu will appear with the option \textit{Investigate feature} (Figure~\ref{fig:overview}). Once this option is selected, the view of the heatmap will change. The layout will be rearranged into two sections: one for the \textit{Investigation features} and below it, a section for \textit{Related features}. 

\begin{itemize}
    \item \textbf{Related features:} This section is divided into \textit{ancestor} and \textit{descendant} features. These are fields that were used to compute the current feature and fields that were computed from the current feature, respectively. This allows the data scientist to see how the changes in that feature correlate with the changes in their related features (\ref{rCompareFeature}). The text input box on the \textit{Investigation features} Y-axis can be used to search and select another feature to add to the view. When there are multiple features in the investigation, the user can choose to see all related features or only the ones common to these features.
\end{itemize}

\subsubsection{Understanding Individual Features}

\begin{figure}
    \centering
	\includegraphics[width=0.5\textwidth]{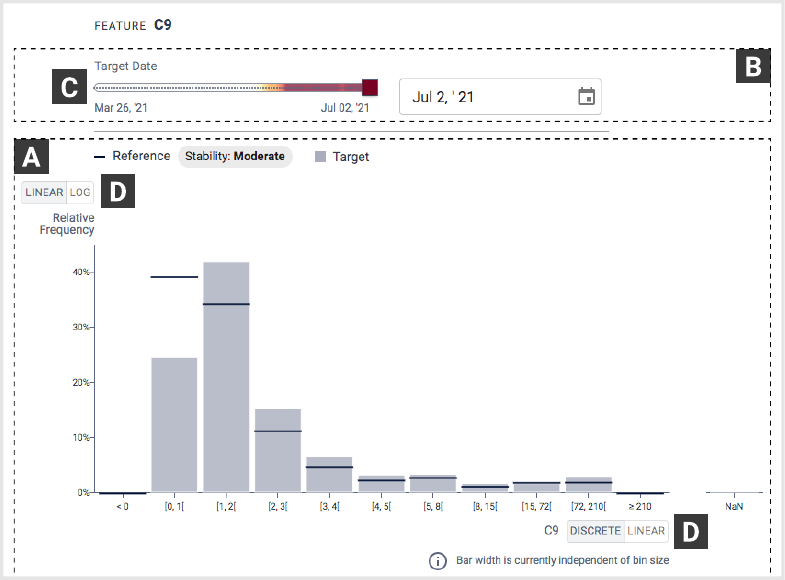}
	\caption{Detail screen. Histogram (A) to visualize how the distribution of a feature has changed. Toolbar (B) with slider (C) to select a target date. The scale of the axes can be changed (D).}
	\label{fig:histogramLayout}
\end{figure}

The histogram method shows how the distribution of a single feature changed over time (\ref{rDistributions}). The main element of the application is the histogram/bar chart (Figure~\ref{fig:histogramLayout}.A), where it is possible to see the relative frequency of a bin (for numerical features) or value (for categorical features) at a target date and compare it with the relative frequency in the reference period. The frequencies at the target date are represented by the light grey bars and at the reference period by horizontal black lines. There is a separate bar on the right of the main histogram that shows the frequency of \textit{NaN} values, in the case of a numeric feature, and the total of missing and new values not seen in the reference period, in the case of a categorical feature.

\begin{itemize}
    \item \textbf{Target date selector:} On top of the histogram, there is a toolbar (Figure~\ref{fig:histogramLayout}.B) where it is possible to select a target date. This can be done using a slider (Figure~\ref{fig:histogramLayout}.C) that covers the time interval in analysis or with a date picker. The slider uses the same color encoding of the heatmap to visualize how much the distributions diverged over time. The date picker is most useful when selecting a specific date of interest.
    \item \textbf{Switchable axis scales:} On the histogram view, the user can use toggle buttons (Figure~\ref{fig:histogramLayout}.D) to change the Y-axis to use either linear (default) or logarithmic scales. For numerical features, it is also possible to change the X-axis to use either a discrete (default) or linear scale. 
    \item \textbf{Brush interaction:} To filter the feature domain on the X-axis, the user can use the brush by clicking and dragging to select a region on the chart area.
\end{itemize}

\subsection{Implementation}



\textit{Data+Shift} is implemented as a Python package. Its API works with data structures used in data science such as pandas DataFrames~\cite{mckinney-proc-scipy-2010,reback2020pandas}. The visualizations are React web applications where charts are built with \texttt{D3}\cite{2011-d3} and \texttt{visx}~\cite{visx} in \texttt{Canvas} and \texttt{SVG}. User-interface components, such as buttons and selections, are from MUI~\cite{mui}. The back-end, responsible for calculating the \textit{p}-values and signaling the alerts is implemented in Python. The visualization methods take as input the result of the evaluation of a dataset and plot the visualizations using IPython's \texttt{display()}. Our implementation approach is similar to \textit{PipelineProfiler}\cite{ono2020pipelineprofiler}. The goal of the design is to integrate \textit{Data-Shift} with the existing workflows of data scientist in JupyterLab (\ref{rWorkflow}).

\section{Expert Feedback}

To validate the suitability of \textit{Data+Shift} for our goals, we ran a second interview with a data scientist at Feedzai who had previously helped us understand design requirements. For this experiment, we considered data on model development for fraud detection at an e-commerce merchant, as the data scientist was already familiar with the use case. We asked the data scientist to use \textit{Data+Shift} to freely explore data distribution shifts during a 45-minute session. We requested him to describe his thought process, following a think-aloud protocol. In this section, we present some of the findings from this session.

When sorting features by \textit{Most alarms} in the \textbf{overview screen}, he immediately noticed five features being alerted from the first day of the analysis. "I wasn't expecting them to be so different. I don't see a reason why it would be this different. This could suggest that there is an issue with data quality." He considered the sorting option \textit{Least sum of \textit{p}-values} to be the most useful to identify drifting features and preemptively detect risks for model performance. The data scientist appreciated the \textbf{investigation screen}. "This is interesting. It makes a lot of sense and helps me understand other features that can be affected by this drift." He pointed out that this view can be useful to detect the source of the shift: "An engineered feature can be calculated from many raw fields, and seeing the \textit{p}-values of its raw ancestors helps me understand which might be causing the drift." When using the \textbf{histogram}, he enjoyed the slider to see the distribution change over time. "I really like the color encoding in the slider. It really helps to understand what is going on with the feature."

The data scientist mentioned that the workflow that took around 20 minutes with \textit{Data+Shift} can take up to one hour without it. "The tool allows me to quickly direct my attention to the source of the problem instead of manually going feature by feature." However, he pointed out that the complete analysis of these shifting features was still necessary. A relevant task that he mentioned is reviewing the data records of each respective drift pattern. The data scientist also hinted that it could be pertinent to explore multidimensional feature drift.

\section{Limitations and Future Work}

During the development of \textit{Data+Shift}, we iterated our solution with the help of data scientists. However, the proposed solution lacks formal user testing. As such, we would like to understand the suitability of this tool with more data scientists to identify potential improvements. Additionally, there is functionality that we did not prioritize for this first version but that we believe can add value to the tool. As examples, we want to implement a more flexible way to truncate Y-axis labels, add visual links between each feature and their related ones, and a mechanism to take stateful screenshots. To make \textit{Data+Shift} more accessible, we also intend to improve support for keyboard navigation and screen readers. Finally, we consider that it is valuable to expand the functionality of \textit{Data+Shift} to support data scientists in a more granular subsequent analysis, exploring ways to visualize individual data records.

\section*{Acknowledgments}

We want to thank Marco O. P. Sampaio for developing the algorithm under \textit{Data+Shift} and sharing the first ideas to visualize it. We are also very grateful to the reviewers for their feedback, helping us shape this manuscript.

\bibliographystyle{eg-alpha-doi}
\bibliography{egbibsample}


\end{document}